\documentclass{article}

\usepackage{microtype}
\usepackage{graphicx}
\usepackage{subcaption}
\usepackage{caption}
\usepackage{booktabs} 

\usepackage{hyperref}

\usepackage{amsmath,amsfonts,bm}

\def\eqref#1{equation~\ref{#1}}

\def\1{\bm{1}}

\DeclareMathAlphabet{\mathsfit}{\encodingdefault}{\sfdefault}{m}{sl}
\SetMathAlphabet{\mathsfit}{bold}{\encodingdefault}{\sfdefault}{bx}{n}

\usepackage{multirow}

\usepackage[accepted]{icml2026}

\usepackage{amsmath}
\usepackage{amssymb}
\usepackage{mathtools}
\usepackage{amsthm}
\usepackage[table]{xcolor}

\usepackage[capitalize,noabbrev]{cleveref}

\theoremstyle{plain}

\theoremstyle{definition}

\theoremstyle{remark}

\usepackage[textsize=tiny]{todonotes}

\icmltitlerunning{NNiT: Width-Agnostic Neural Network Generation with Structurally Aligned Weight Spaces}

\begin{document}

\twocolumn[
  \icmltitle{NNiT: Width-Agnostic Neural Network Generation with Structurally Aligned Weight Spaces}



  \icmlsetsymbol{equal}{*}

  \begin{icmlauthorlist}
    \icmlauthor{Jiwoo Kim}{Dukeece}
    \icmlauthor{Swarajh Mehta}{Dukecs}
    \icmlauthor{Hao-Lun Hsu}{Dukecs}
    \icmlauthor{Hyunwoo Ryu}{Miteecs}
    \icmlauthor{Yudong Liu}{Dukeece}
    \icmlauthor{Miroslav Pajic}{Dukeece}
  \end{icmlauthorlist}

  \icmlaffiliation{Dukeece}{Department of Electrical and Computer Engineering, Duke University, NC, USA}
  \icmlaffiliation{Dukecs}{Department of Computer Science, Duke University, NC, USA}
  \icmlaffiliation{Miteecs}{Computer Science \& Artificial Intelligence Laboratory, Massachusetts Institute of Technology, MA, USA}

  \icmlcorrespondingauthor{Jiwoo Kim}{jiwoo.kim@duke.edu}
  \icmlcorrespondingauthor{Miroslav Pajic}{miroslav.pajic@duke.edu}

  \icmlkeywords{Diffusion Model, Weight Space Learning, Neural Network Diffusion}

  \vskip 0.3in
  
]



\printAffiliationsAndNotice{}  


\begin{abstract}
    Generative modeling of neural network parameters is often tied to architectures because standard parameter representations rely on known weight-matrix dimensions. Generation is further complicated by permutation symmetries that allow networks to model similar input-output functions while having widely different, unaligned parameterizations. In this work, we introduce Neural Network Diffusion Transformers (NNiTs), which generate weights in a width-agnostic manner by tokenizing weight matrices into patches and modeling them as locally structured fields. We establish that Graph HyperNetworks (GHNs) with a convolutional neural network (CNN) decoder structurally align the weight space, creating the local correlation necessary for patch-based processing. Focusing on Multilayer Perceptrons (MLPs), where permutation symmetry is especially apparent, NNiTs generate fully functional networks across a range of architectures. Our approach jointly models discrete architecture tokens and continuous weight patches within a single sequence model. On ManiSkill3 robotics tasks, NNiT achieves $>85\%$ success on architecture topologies unseen during training, while baseline approaches fail to generalize; the same pipeline also generalizes to MNIST classification beyond the robotic control setting.
\end{abstract}



\section{Introduction}
\begin{figure*}
    \centering
    \includegraphics[width=0.92\linewidth]{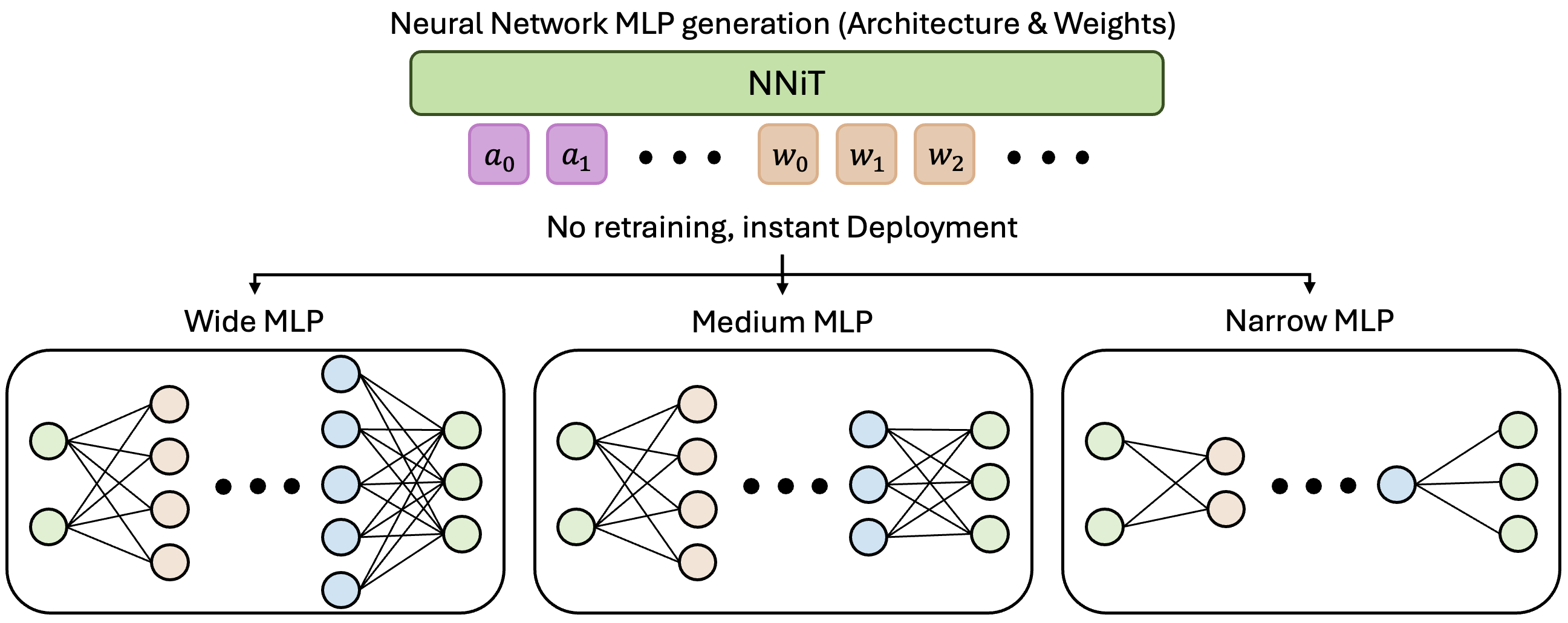}
    \caption{\textbf{Width-Agnostic Synthesis via Multimodal Tokenization}. Unlike existing models, the NNiT decouples functional logic from fixed matrix dimensions, allowing the zero-shot synthesis of optimal weights for architectural topologies entirely unseen during training.}
    \label{fig:intro}
\end{figure*}

Diffusion Transformers have demonstrated scalability across various modalities, such as computer vision image generation~\cite{Peebles2022DiT, bao2022all, xie2025sana, xie2024sana, gao2023masked}, video generation~\cite{liu2025javisdit,ho2022imagenvideohighdefinition, Menapace_2024_CVPR,chen2025sana, ma2025latte,gupta2023photorealisticvideogenerationdiffusion}, 3D generation ~\cite{mo2023dit3d,xiang2024structured,shuang2025direct3d}, and protein or molecule design~\cite{wu2022proteinstructuregenerationfolding,joshi2025allatomdiffusiontransformersunified,liu2024graphdiffusiontransformersmulticonditional,li2025proteinaeproteindiffusionautoencoders}. Recently, this generative paradigm has been applied to parameter synthesis, where models directly generate weights for functional neural networks~\cite{liang2024make, soro2024diffusionbasedneuralnetworkweights, Erkoc_2023_ICCV}. These approaches aim to sample full neural networks from a learned distribution, bypassing the computational cost of traditional training.

A central challenge in neural parameter synthesis is the permutation symmetry of network weights~\cite{navon2023equivariant, zhao2025symmetry}, where many distinct parameterizations correspond to the same function, so adjacent weights are spatially uncorrelated. To address this challenge, recent approaches generate weights in a latent embedding~\cite{soro2024diffusionbasedneuralnetworkweights, saragih2025flowlearnflowmatching}, or apply explicit canonicalization methods~\cite{ainsworth2023gitrebasinmergingmodels} before flattening parameters into 1D vectors~\cite{schuerholt2024sane}. While these approaches can successfully transfer across varying depths, they remain fragile to changes in width. Once a weight matrix is collapsed into a fixed-dimensional vector, the generative prior becomes coupled to the matrix size seen during training. Even with alignment, changing the layer width changes token dimensionality and disrupts learned correspondence, preventing generalization to geometries unseen during training.


Consequently, to resolve this limitation, we rethink the role of Graph HyperNetworks (GHNs)~\cite{zhang2020graphhypernetworksneuralarchitecture, knyazev2021parameter} and apply them as both a data source generator and a mechanism for aligning weight space. GHNs propagate information over the architecture graph and generate layer parameters in fixed order, anchored at an input node with task features. In our implementation of the GHN, we use a CNN decoder, which imposes an explicit locality bias in the weight-space. This results in weights with consistent local spatial correlations across the population. On the other hand, the standard Stochastic Gradient Descent (SGD) typically produces functionally equivalent solutions with arbitrary permutations resulting in unaligned parameter space. The GHN-induced parameter distribution thus provides the local structure needed for patch-based tokenization and width-agnostic transfer.

Building on this alignment, we introduce 
the Neural Network Diffusion Transformer (NNiT), illustrated in Figure~\ref{fig:intro},
which formulates neural network synthesis as a single multimodal sequence modeling task. NNiT tokenizes aligned weight tensors into $p \times p$ patches, replacing global vectors with spatially correlated tokens. This representation makes generation width-agnostic---widening a layer just corresponds to generating additional patches without changing the tokenization scheme.~An NNiT therefore learns a \emph{joint distribution over architectures and parameters}, allowing generation of unseen architecture-weight pairs and the zero-shot synthesis of weights for user-specified structures.

We evaluate NNiT on ManiSkill3~\cite{taomaniskill3} using Multilayer Perceptron (MLP) policies for robotic control. Robotic manipulation is a strict evaluation metric where small weight errors may cause task failures, unlike vision tasks where artifacts could be tolerated. Moreover, generating robotic policies is an underexplored space where enabling joint weight and architecture generation opens new avenues for research, such as meta-learning across tasks or optimizing networks for specific hardware constraints for sim to real deployment. Since this domain naturally favors MLPs, it serves as an ideal benchmark for evaluating structural generalization. 
Specifically, we show that NNiTs maintain $>85\%$ success rates on architectural widths unseen during training, while baselines~\cite{schuerholt2024sane, soro2024diffusionbasedneuralnetworkweights} fail.
Beyond robotic control, we show that the same pipeline also generalizes to unseen architectures on MNIST classification.

To summarize, our primary contributions are as follows:
\begin{itemize}
    \item We establish that 
    GHNs align the weight space, reducing permutation-induced variability and enabling coordinate-based parameter field for tokenization.
    \item We introduce patch tokenization for weights, making generation width-agnostic and enabling zero-shot synthesis for unseen architecture topologies.
    \item We introduce NNiT, a multimodal diffusion transformer that jointly models architectures and weights for both joint generation and conditional weight synthesis.
\end{itemize}

\section{Related Work}

\subsection{Generative Neural Network Weight Models}
Parameter synthesis is increasingly formulated as a direct generative modeling problem, where neural weights are treated as structured data modalities~\cite{wang2025recurrent, li2025texttomodeltextconditionedneuralnetwork, saragih2025flowlearnflowmatching, Erkoc_2023_ICCV}. Recent approaches have successfully adapted diffusion~\cite{ho2020denoisingdiffusionprobabilisticmodels,song2022denoisingdiffusionimplicitmodels} and flow matching models~\cite{lipman2023flowmatchinggenerativemodeling} to approximate these complex parameter distributions. However, a fundamental constraint in this domain is the high dimensionality and inherent permutation symmetry of weight spaces, where arbitrary neuron ordering obscures geometric structure.

To resolve this ambiguity, prior works have employed Variational Autoencoders (VAEs) to compress weights into permutation-invariant latent codes~\cite{saragih2025flowlearnflowmatching, soro2024diffusionbasedneuralnetworkweights}. Alternatively, methods such as SANE~\cite{schuerholt2024sane} utilize explicit canonicalization methods to align the weights into a shared coordinate frame. In this vein, recent studies~\cite{ainsworth2023gitrebasinmergingmodels,rinaldi2025updatetransformerlatestrelease} validate that offline optimization can effectively collapse weight permutation symmetries. Notably, these approaches have demonstrated transferability across diverse architectural depths, successfully synthesizing parameters for Convolutional Neural Networks (CNNs)~\cite{schuerholt2024sane, soro2024diffusionbasedneuralnetworkweights}; yet, they typically remain constrained to fixed-width architectures.

\subsection{Diffusion Models}
Diffusion probabilistic models have established themselves as the standard backbone for generative tasks, demonstrating exceptional scalability across image~\cite{Peebles2022DiT, bao2022all,ma2024sit,wang2025native,tian2024udits} and video generation~\cite{liu2025javisdit, ruan2022mmdiffusion,kim2024a}, as well as robotics~\cite{kim2023robotic, chi2023diffusionpolicy, ryu2023diffusion, chang2023denoising, qi2025ecdiffuser}. Diffusion Transformers (DiTs) advanced this paradigm by utilizing patch-based tokenization to process data as sequences. This structural innovation facilitates multimodal training by projecting diverse data types—such as audio, video, and text—into a unified representation space~\cite{liu2025javisdit, ruan2022mmdiffusion,kim2024a, shin2025exploringmmdit,kim24openvla,zhou2024transfusion,gao2024lumin-t2x}. Most relevant to our framework, recent Flexible Diffusion Transformers (FiTs)~\cite{Lu2024FiT, wang2024fitv2} have introduced mechanisms to process variable-resolution images. We leverage these primitives to decouple parameter synthesis from fixed tensor dimensions, treating neural weights as width-agnostic fields.

\section{Preliminaries and Problem Formulation}

\begin{figure*}[t]
    \centering
    \includegraphics[width=1.0\linewidth]{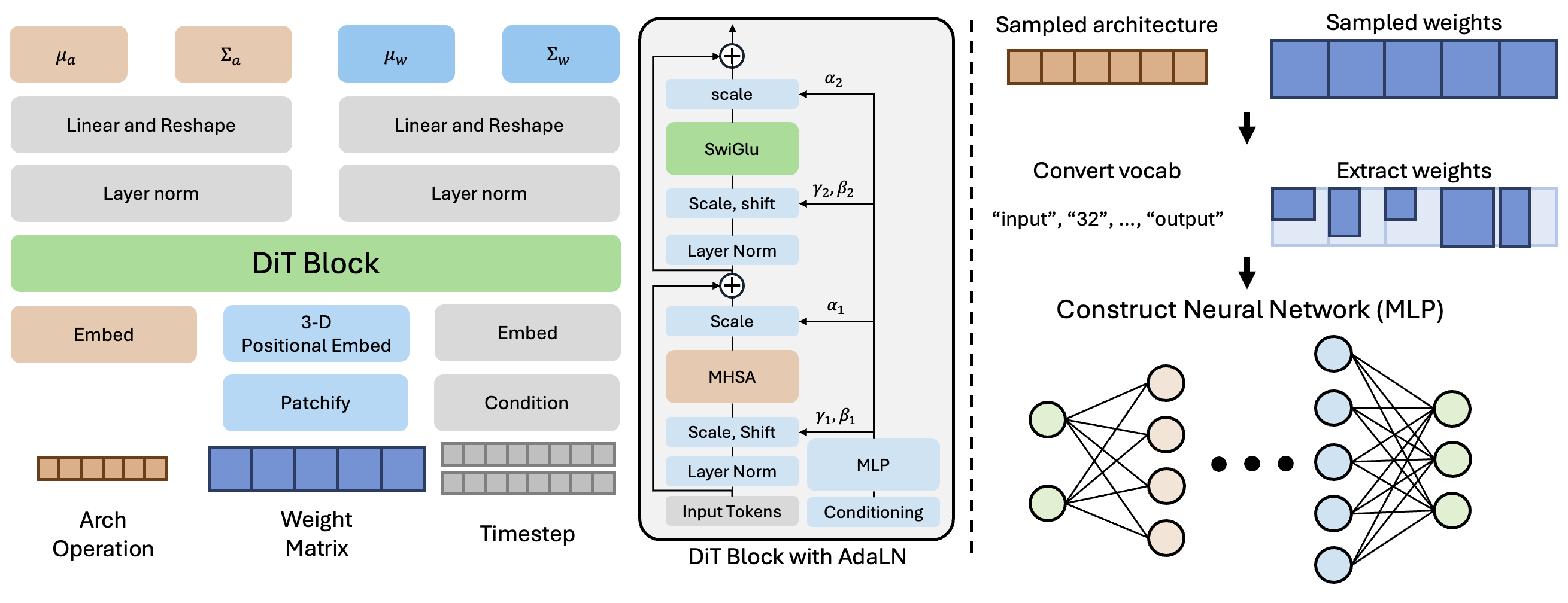}
    \caption{\textbf{NNiT Framework Overview.} \textbf{Left: Unified Generative Architecture.} We formulate neural synthesis as a multimodal sequence task. Discrete architecture tokens (orange) and continuous weight matrices (blue) are unified into a single sequence, with weights processed as spatially correlated patches. A Diffusion Transformer (DiT) models the joint distribution using per-modality timestep conditioning ($\mu_a, \Sigma_a, \mu_w, \Sigma_w$) via the Mixture of Noise Levels (MoNL) framework, enabling both co-design $p(\mathbf{a},\mathbf{w})$ and conditional synthesis $p(\mathbf{w}|\mathbf{a})$. \textbf{Right: Deployment Pipeline.} During inference, sampled architecture tokens are decoded into layer widths. The generated weight tensors are then extracted to match these target dimensions, assembling a directly executable MLP.}
    \label{fig:method}
\end{figure*}

\subsection{Permutation Symmetry}
\label{sec:permutation_symmetry}
For an $L$-layer MLP, the parameters $\mathbf{w} = \{W_l, b_l\}_{l=1}^{L}$ are subject to inherent permutation symmetries \cite{hecht1990algebraic}.
The loss function is invariant under transformations by permutation matrices $P_l \in \mathbb{R}^{d_{l+1} \times d_{l+1}}$ (where $P_l^T P_l = I$) applied to the hidden units of layer $l$.
For any set of permutations $\pi = \{P_l\}_{1}^{L}$, the input-output function of the network remains 
the same because the activations satisfy: 
\begin{equation}
    z_{l+1} = P^T \sigma(P W_l z_l + P b_l),
\end{equation}
where $\sigma$ is the non-linear activation function. This demonstrates that permuting the output of layer $l$ by $P$, while simultaneously applying the inverse permutation $P^T$ to the input of layer $l+1$, leaves the resulting activations functionally invariant~\cite{ainsworth2023gitrebasinmergingmodels}. 

\subsection{Graph HyperNetworks (GHNs)}
\label{sec:pre_ghn}
A 
GHN~\cite{zhang2020graphhypernetworksneuralarchitecture, knyazev2021parameter} acts as a deterministic meta-model $\Phi_\phi$ parameterized by $\phi$. Instead of optimizing weights directly via SGD, a GHN predicts the parameters $\textbf{w}$ for a given neural architecture $\textbf{a}$, represented as a computational graph of parameterized operations $\mathcal{G} = (\mathcal{V}, \mathcal{E})$. Here, $\mathcal{V}$ represents the set of parameterized operations, and $\mathcal{E}$ represents the connectivity between them.

The process begins by initializing the node embeddings  $\mathbf{H}^{(0)} = \{\mathbf{h}_v^{(0)}\}_{v \in \mathcal{V}}$ based on the properties of each node, such as the layer shape and operation type. These embeddings are updated iteratively via a Graph Neural Network (GNN). At step $t$, the embedding for node $v$ is updated aggregating messages from these neighbors:
\begin{equation} 
\mathbf{h}_v^{(t+1)} = U_{\theta}\left(\mathbf{h}_v^{(t)}, \sum\nolimits_{u \in \mathcal{N}(v)} M_{\psi}(\mathbf{h}_u^{(t)})\right),
\end{equation}

where $M_{\psi}(\cdot)$ is a message function that projects neighbor states into the message space, $U_{\theta}(\cdot)$ is a node update function that integrates the aggregated neighborhood information with the node's current state, and $\mathcal{N}(v)$ denotes the set of incoming neighbors for node $v$. After T steps of propagation, a shared decoder network $D_\phi$ projects the final node embedding into the weight parameters for that specific layer  $\mathbf{w}_v = D_\phi(\mathbf{h}_v^{(T)})$.

\subsection{Problem Objective}
We formulate our generative model as learning a joint distribution $p_\theta(\mathbf{a}, \mathbf{w})$ over the space of topologies $\mathbf{a} \in \mathcal{A}$ and parameters $\mathbf{w} \in \mathcal{W}$. A functional network is defined by the tuple $\tau = (\mathbf{a}, \mathbf{w})$.

\textbf{Objective.} Our primary goal is to achieve width-agnostic neural network MLP generation, enabling the model to generalize to unseen topologies not present in the training set. Existing generative approaches typically flatten weights into fixed-dimensional vectors, rigidly coupling the generative prior to specific architectures, or similar architectures with different depths~\cite{schuerholt2024sane,soro2024diffusionbasedneuralnetworkweights}.

\textbf{Problem.} 
Designing methods for such width-agnostic synthesis is fundamentally difficult due to the permutation symmetry and lack of geometry described in Section~\ref{sec:permutation_symmetry}. In standard neural networks trained via SGD, adjacent weights in a matrix are uncorrelated due to the arbitrary ordering of neurons, breaking the core structure required for spatial generation.

\textbf{Key Insight and Proposed Approach.} To address this challenge, we demonstrate that 
GHNs with a CNN decoder inherently generate a structurally aligned weight space. We observe that, unlike SGD, which produces unstructured weight matrices, GHNs collapse permutation symmetries into a consistent topological structure. Leveraging this finding, we propose \emph{treating neural network weights not as independent vectors, but as a continuous spatial field}. This structural alignment allows us to employ image-based generative backbones to model the conditional likelihood $p_\theta(\mathbf{w} | \mathbf{a})$ across varying topologies as well as joint multimodal generation of $p_\theta(\mathbf{a},\mathbf{w})$.

\section{Structural Alignment via Graph-Conditioned CNN Decoder}
Our NNiT framework builds on the empirical observation that GHNs, when paired with a CNN decoder, produce weight tensors that are consistently organized and exhibit reliable local correlations. This effectively narrows the training distribution seen by our downstream generative model; instead of learning over the full, permutation-ambiguous space of MLP parameters, it learns over a more structurally coherent subset of MLPs induced by the GHN generator. This addresses the two main obstacles to field-based weight generation: permutation ambiguity and the lack of stable spatial~structure.

We attribute the reduction in permutation-induced variability to the GHN's parameterization mechanism. The GHN computes embeddings $\mathbf{h}_v$ for each node by propagating information over the architecture graph. The input node $\mathcal{V}_{in}$ is anchored by task-specific features, providing a consistent reference across samples. A shared decoder $D_\phi(\cdot)$ then maps these embeddings to the weight tensors of each layer.

Crucially, we implement $D_\phi$ as a CNN decoder. The CNN decoder maps a compact node embedding to a full parameter tensor, $D_\phi:\mathbb{R}^d \rightarrow \mathbb{R}^N$ with $d \ll N$. It introduces an explicit locality bias on the generated weights where nearby indices are produced from shared latent features. This yields repeatable spatial patterns like the vertical banding structures seen in Figure~\ref{fig:ghn_vs_mlp_heatmap}. We also see consistent local correlations across seeds. 

In contrast, SGD-trained MLPs have parameterizations with no stable alignment. This GHN generator thus produces a weight distribution with reliable local structure, which is the prerequisite needed for patch-based tokenization and width-agnostic generation using NNiT.

\section{Neural Network Diffusion Transformers}

Leveraging the structurally aligned weight-space induced by the GHN, we introduce the NNiT, a multimodal framework that unifies neural architecture search and parameter generation into a single sequence modeling task (Figure~\ref{fig:method}).

\subsection{Unified Neural Network Representation}
\label{sec:embedding}

We define a joint embedding space that processes discrete architecture tokens and continuous weight patches simultaneously.

\subsubsection{Architecture as Discrete Tokens}

We formulate the architecture $\mathbf{a} \in \mathcal{V}^S$ as a sequence of discrete tokens, where $\mathcal{V}$ is the vocabulary of layer widths and $S$ denotes the network depth. In contrast to approaches that utilize adjacency matrices to model arbitrary branching topologies~\cite{an2023diffusionnag}, we formulate the architecture as a discrete sequence. This representation is sufficient for MLP policies where full layer connectivity is implicit in the sequence order. Accordingly, each token $a_i$ serves as a discrete index mapping to a specific neuron count within $\mathcal{V}$. These indices are projected to the model's hidden dimension $d$ via a learnable embedding table $E_a$, resulting in the dense vector sequence $\mathbf{z}_a \in \mathbb{R}^{S \times d}$.

\subsubsection{Weights as Patchified Continuous Tensor}
To achieve width-agnostic generation, we leverage the structural alignment of the structurally aligned weight spaces to frame the weights as a continuous 4D tensor. For a specific layer $l$, we concatenate the weight matrix and bias vector to form a parameter block $B_l \in \mathbb{R}^{n_l \times (n_{l+1} + 1)}$. To accommodate diverse topologies within a single generative model, we standardize these blocks by padding them onto a maximal coordinate grid $H \times W$, where $H=M$, $W=M+p$, with $M$ being an arbitrary maximum supported width following~\cite{Lu2024FiT}. The complete set of network parameters is thus represented as a 4D tensor $\tilde{\mathbf{w}} \in \mathbb{R}^{(S-1) \times 1 \times H \times W}$.

We then decompose this tensor into non-overlapping $p \times p$ patches locally within each layer. This results in a sequence of flattened patches $\mathbf{w}_{patch} \in \mathbb{R}^{(S-1) \cdot N \times p^2}$, where $N = \frac{H \cdot W}{p^2}$ is the number of patches per layer grid. These patches are linearly projected to the model dimension $d$ to form the continuous embedding sequence $\mathbf{z}_w \in \mathbb{R}^{L_{seq} \times d}$.

The final input to the transformer is the unified sequence $\mathbf{z} = [\mathbf{z}_a; \mathbf{z}_w]$, allowing the self-attention mechanism to model the joint dependencies between the discrete topology tokens and the functional weight patches.

\subsection{Transformer Backbone}
NNiT processes the unified sequence $\mathbf{z}$ using a standard Diffusion Transformer (DiT) backbone~\cite{Peebles2022DiT}.~Each block consists of multi-head self-attention and a feed-forward network utilizing SwiGLU activations like  LLaMA~\cite{touvron2023llama}.

To handle the heterogeneous nature of the input, conditioning is implemented via Adaptive Layer Norm (AdaLN-Zero) with dual-timestep embeddings. We employ distinct timestep embedding models for the architecture ($t_a$) and weights ($t_w$). These embeddings are concatenated to dynamically regulate the scale and shift parameters of the layers, allowing the network to modulate its processing based on the noise level of each modality.

\subsection{Training with Mixture of Noise Levels (MoNL)}
\label{sec:monl}

Finally, we adopt the Mixture of Noise Levels (MoNL) framework~\cite{kim2024a} to unify architecture generation and parameter synthesis; we provide more details in Appendix~\ref{app:monl}. During training, we stochastically sample between two noise scheduling modes for each~batch:
\begin{itemize}
    \item Joint Generation Mode ($t_a = t_w > 0$): Both architecture and weights are diffused to the same timestep $t_{\text{ref}}$. The model learns the joint distribution $p(\mathbf{a}, \mathbf{w})$, enabling it to propose novel architecture-weight pairs from scratch.
    \item Conditional Synthesis Mode ($t_a = 0, t_w > 0$): Architecture tokens remain noise-free ($t_a=0$) while weights are diffused to $t_{\text{ref}}$. This forces the model to learn the conditional distribution $p(\mathbf{w}_{t-1} | \mathbf{w}_t, \mathbf{a}_0)$, effectively teaching it to synthesize valid weights for a fixed topology.
\end{itemize}

To optimize this joint distribution, we minimize a composite loss function. Given clean data $(\mathbf{a}_0, \mathbf{w}_0)$ and independent Gaussian noise $\boldsymbol{\epsilon}_a, \boldsymbol{\epsilon}_w \sim \mathcal{N}(0, \mathbf{I})$, the total objective combines the standard noise prediction error with a variational lower bound term $\mathcal{L}_{\text{vb}}$ to learn the covariance $\Sigma_\theta$ -- i.e.,
\begin{equation}
    \mathcal{L}_{\text{total}} = \mathbb{E}_{t_a, t_w, \boldsymbol{\epsilon}} \Big[ \| \boldsymbol{\epsilon}_\theta^a - \boldsymbol{\epsilon}_a \|^2 + \| \boldsymbol{\epsilon}_\theta^w - \boldsymbol{\epsilon}_w \|^2 \Big] + \mathcal{L}_{\text{vb}},
\end{equation}
where $\boldsymbol{\epsilon}_\theta^a$ and $\boldsymbol{\epsilon}_\theta^w$ denote the model's noise predictions for the architecture and weight modalities, respectively.

\subsection{Deployment and Synthesis}
\label{seq:deploy}
Inference adapts to the presence of architectural constraints. If a target architecture is provided, the model samples $\mathbf{w} \sim p(\mathbf{w}|\mathbf{a})$. Otherwise, it samples from the joint distribution $(\mathbf{a},\mathbf{w}) \sim p(\mathbf{a}, \mathbf{w})$.

The final network assembly is a deterministic projection driven by the architecture tokens. The sequence  $\hat{\mathbf{a}}$ is decoded into integer widths $[n_1, \ldots, n_S]$ where $n_1, ..., n_S \in V$. These dimensions serve as a cropping mask for the generated weight tensor. We extract the valid submatrices $\mathbf{w}_l \in \mathbb{R}^{n_l \times (n_{l+1}+1)}$ from the maximal grid and discard the padding. 

\section{Experimental Evaluation}

\begin{figure*}
    \centering
    \includegraphics[width=\linewidth]{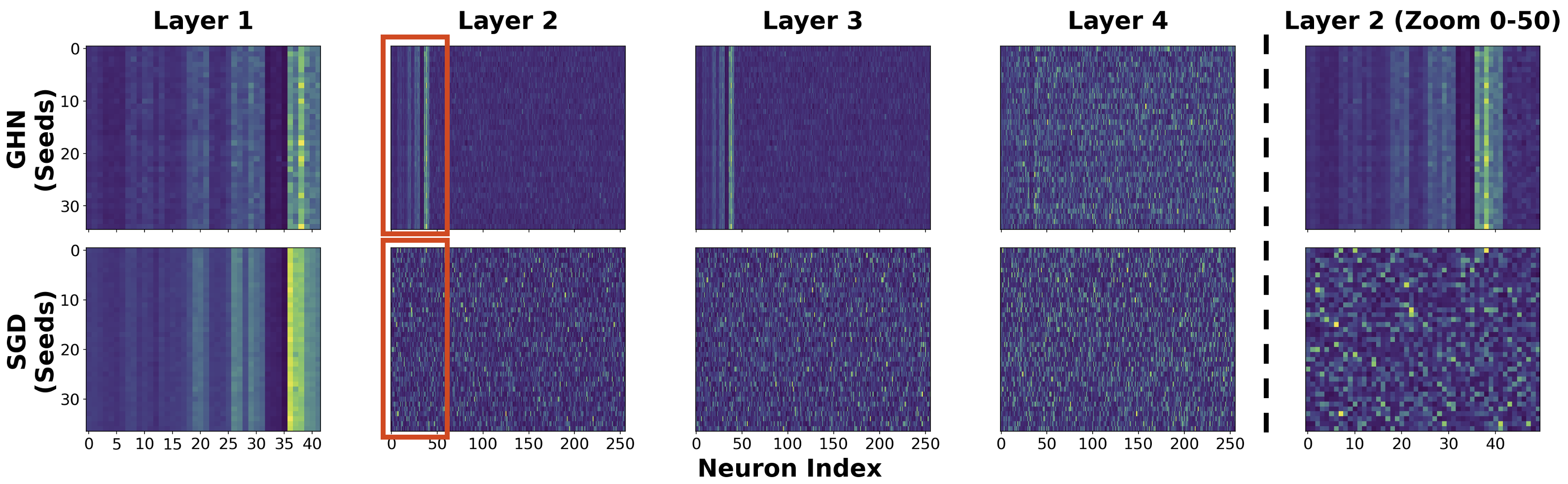}
    \caption{\textbf{Visualizing Structural Alignment and Induced Geometry.} Comparison of the weight magnitude profiles across 35 independent seeds. \textbf{Top (GHN):} The consistent alignment across the seeds demonstrates that the GHN \textbf{1) successfully spatially aligns the weight spaces}, effectively resolving the permutation ambiguity inherent in neural networks. Moreover, the visible structural banding indicates that the GHN \textbf{2) imposes meaningful geometric structure} (spatial correlation), transforming independent parameters into a spatially aligned space. \textbf{Bottom (SGD):} In contrast, SGD weights exhibit unstructured noise due to arbitrary permutations, lacking both structural alignment and local spatial geometry.}
    \label{fig:ghn_vs_mlp_heatmap}
\end{figure*}

We aim to demonstrate that the NNiT learns a width-agnostic representation of neural weights, enabling transferability across diverse architectures. In particular, 
we consider the following research questions:

\begin{itemize}
    \item \textbf{Does the GHN backbone effectively reduce permutation ambiguity?} We investigate if GHNs induce the necessary structural alignment to treat weights as local patches (Section~\ref{sec:ghn_vs_sgd}).
    
    \item \textbf{Is the NNiT more effective than baselines in zero-shot transfer?} We evaluate whether our patch-based tokenization outperforms existing baseline models (SANE~\cite{schuerholt2024sane} and D2NWG~\cite{soro2024diffusionbasedneuralnetworkweights}) when synthesizing weights for architectural topologies with diverse widths (Section~\ref{sec:width_transfer_exp}).
    
    \item \textbf{Can NNiT perform multimodal joint generation?} We assess the NNiT's ability to model the joint distribution $p(\mathbf{a}, \mathbf{w})$, spontaneously generating both diverse network topologies and their functional parameters (Section~\ref{sec:joint_synthesis}).
\end{itemize}

\subsection{Empirical Validation of Structural Alignment}
\label{sec:ghn_vs_sgd}

\begin{table}[t]
    \centering
    \caption{\textbf{Alignment vs. Performance.} Evaluation of 35 independent seeds. Both methods achieve optimal performance, yet GHNs induce the structural alignment necessary for tokenization without sacrificing weight diversity, confirming the absence of mode collapse.}
    \label{tab:ghn_sgd_metrics}
    \begin{tabular}{lcc}
        \toprule
        \textbf{Metric} & \textbf{GHN} & \textbf{SGD} \\
        \midrule
        \multicolumn{3}{l}{\textit{Policy Performance}} \\
        Mean Return & $39.35$ & $39.49$ \\
        Mean Success Rate & $99.0\%$ & $98.9\% $ \\
        \midrule
        \multicolumn{3}{l}{\textit{Weight Diversity (Nearest Neighbor)}} \\
         $L_2$ Distance ($\uparrow$) & $54.66$ & $41.73$\\
        Cosine Similarity ($\downarrow$) & $0.07$ & $0.02$ \\
        \bottomrule
    \end{tabular}
\end{table}

We first test the hypothesis that GHNs induce the structural conditions necessary for patch-based generation. Using \texttt{PickCube-v1} as a representative environment, we compare weight-magnitude profiles from 35 independently sampled GHN-generated policies and 35 standard SGD-trained networks. Both sets use the same MLP architecture, $[\text{input}, 256, 256, 256, \text{output}]$.

As shown in Table~\ref{tab:ghn_sgd_metrics}, both methods achieve expert performance ($>99\%$ success). Figure~\ref{fig:ghn_vs_mlp_heatmap} nevertheless shows a clear structural difference. Because the input feature space is fixed across seeds, both GHN and SGD exhibit pronounced vertical banding in the input projection layer. Beyond the first layer, however, SGD-trained weights display little consistent spatial structure across seeds.

On the other hand, GHN-generated weight-magnitude heatmaps retain vertically banded, locally correlated structure across all layers. This suggests that generating parameters through the shared, graph-conditioned decoder selects MLPs with aligned weight spaces. 

Crucially, this alignment is not a result of mode collapse. The pairwise nearest neighbor analysis in Table~\ref{tab:ghn_sgd_metrics} reveals that GHN policies maintain high weight diversity with a mean $L_2$ distance of 54.66 and a small mean cosine similarity of 0.07.~Additional per-architecture heatmaps and diversity metrics in Appendix~\ref{app:data_dist} further support this finding. We also provide additional evidence in Appendix~\ref{app:alignment_ablation}, where a centered kernel alignment (CKA) analysis confirms the absence of mode collapse and a permutation ablation establishes that the NNiT depends on this alignment.

\begin{table*}[t]
\centering
\caption{\textbf{Architecture-conditioned weight generation ($p(\mathbf{w}|\mathbf{a})$).} We compare NNiT against baseline methods on both seen and \textbf{unseen (zero-shot)} architectural configurations. Reporting the top-10 performance metric, we observe that while the gap is negligible on training configurations, performance diverges sharply on unseen topologies. Baselines degrade significantly, while NNiT maintains robust performance, validating its effectiveness as a width-agnostic neural network generator.}
\label{tab:conditional_generation_results}
\begin{tabular}{lccccccc}
\toprule
 & & \multicolumn{2}{c}{PickCube-v1} & \multicolumn{2}{c}{PushCube-v1} & \multicolumn{2}{c}{StackCubeEasy-v1} \\
\cmidrule(lr){3-4} \cmidrule(lr){5-6} \cmidrule(lr){7-8}
 & Architecture & Return & Success & Return & Success & Return & Success \\
\midrule
Dataset & & $39.37 \pm 0.55$ & $98\%$ & $42.02 \pm 0.4$ & $100\%$ & $39.53 \pm 0.78$ & $94\%$ \\
\midrule
SANE & \multirow{3}{*}{Seen} & $3.81 \pm 0.29$ & $1\%$ & $6.90 \pm 0.25$ & $6\%$ & $4.64 \pm 0.16$ & $0\%$ \\
D2NWG & & $38.59 \pm 1.35$ & $98\%$ & $42.28 \pm 0.05$ & $100\%$ & $37.65 \pm 0.69$ & $90\%$ \\

\textbf{NNiT (Ours)} & & $38.77 \pm 0.41$ & $98\%$ & $42.10 \pm 0.16$ & $100\%$ & $38.38 \pm 0.51$ & $91\%$ \\
\midrule
\midrule
\rowcolor{gray!10}
SANE &  & $3.65 \pm 0.41$ & $1\%$& $5.89 \pm 0.32$ & $2\%$ & $3.71 \pm 0.13$ & $0\%$ \\
\rowcolor{gray!10}
D2NWG & & $28.31 \pm 4.53$ & $59\%$ & $37.35 \pm 1.63$ & $89\%$ & $27.18 \pm 2.35$ & $42\%$ \\

\rowcolor{gray!20}
\textbf{NNiT (Ours)} & \multirow{-3}{*}{\textbf{Unseen}}& $\mathbf{38.70 \pm 0.36}$ & $\mathbf{99\%}$ & $\mathbf{42.00 \pm 0.19}$ & $\mathbf{100\%}$ & $\mathbf{36.24 \pm 1.33}$ & $\mathbf{86\%}$ \\
\bottomrule

\end{tabular}
\end{table*}

\begin{table*}[!t]
\centering
\caption{\textbf{Multimodal Joint Synthesis ($p(\mathbf{a},\mathbf{w})$).} We evaluate NNiT's capacity to spontaneously synthesize complete functional policies. As no baselines support joint synthesis, NNiT is presented in isolation. The results demonstrate near-perfect success rates ($99\% \text{--} 100\%$) across PickCube and PushCube and $90\%$ for StackCubeEasy. This confirms that the model can generate high-performing policies without a fixed architectural prompt.}
\label{tab:joint_generation_results}
\begin{tabular}{lccccccc}
\toprule
 & \multicolumn{2}{c}{PickCube-v1} & \multicolumn{2}{c}{PushCube-v1} & \multicolumn{2}{c}{StackCubeEasy-v1} \\
\cmidrule(lr){2-3} \cmidrule(lr){4-5} \cmidrule(lr){6-7}
 & Return & Success & Return & Success & Return & Success & \\
\midrule
NNiT(Ours)  & $38.85 \pm 0.32$ & $99\%$ & $41.99 \pm 0.10$ & $100\%$ & $37.57 \pm 0.85$ & $90\%$  \\
\bottomrule
\end{tabular}
\end{table*}

\subsection{Dataset \& Experimental Design}
\label{sec:dataset_setup}
To evaluate the neural network generation, we constructed a dataset of 4-hidden-layer MLP policies with layer widths sampled from $\mathcal{V} \in \{\text{input},16, 32, 64,\text{output}\}$. We partitioned $72$ distinct topological configurations into $64$ training architectures and $8$ held-out configurations to test zero-shot generalization to unseen structures. We trained $128$ independent GHNs, extracting the top-$100$ performing weight configurations per training architecture. This yielded a final corpus of $6,400$ expert policies for training our NNiT model. Additional details are provided in Appendix~\ref{app:data_gen},  with full training hyperparameters summarized in Appendix~\ref{app:impl_details}.

\subsection{Evaluation Protocol}
We assess the generation quality by directly deploying synthesized policies into the ManiSkill3 environment, evaluating each over 50 episodes to measure control fidelity. Following the standard protocols~\cite{liang2024make, schuerholt2024sane, soro2024diffusionbasedneuralnetworkweights}, we report the performance of the top-10 policies selected from 100 generated samples, simulating an offline validation phase before deployment. To benchmark width-agnostic synthesis methods, we compare NNiT against D2NWG~\cite{soro2024diffusionbasedneuralnetworkweights} and SANE~\cite{schuerholt2024sane}, trained on the same expert dataset. As no prior methods target zero-shot architectural generalization, we adapted both baselines to this setting and evaluated them without test-time optimization.

\subsection{Zero-Shot Width Transferability}
\label{sec:width_transfer_exp}
Table~\ref{tab:conditional_generation_results} presents the definitive test of our framework regarding zero-shot conditional generation. We evaluate the model's ability to synthesize weights for architectures with different topologies (i.e., $p(\mathbf{w}|\mathbf{a})$).

On seen architectures, both the NNiT and the D2NWG~\cite{soro2024diffusionbasedneuralnetworkweights} baseline achieve near perfect performance for all three tasks. These results suggest that the D2NWG~\cite{soro2024diffusionbasedneuralnetworkweights} effectively captures the training distribution via explicit architecture conditioning. We also note that part of the D2NWG's strong performance on seen data is 
due to the GHN pre-aligning the MLP weight spaces, making them more conducive to downstream weight-space learning.

However, such conditioning rigidly couples the generation process to specific topologies because the D2NWG 
relies on fixed zero-padded vectorization schemes lacking explicit architectural awareness. This limitation impairs transferability to unseen structures and causes success rates on \texttt{PickCube-v1} and \texttt{StackCubeEasy-v1} to drop to $59\%$ and $42\%$. While the D2NWG retains $89\%$ success on \texttt{PushCube-v1}, we attribute this to task simplicity, tolerating higher precision errors rather than true structural generalization. 

In contrast, SANE~\cite{schuerholt2024sane} fails to produce functional policies across both seen and unseen topologies. We attribute this failure to the reliance on global positional embeddings for the sequence order encoding, as this formulation becomes unstable when trained on a mixture of diverse network widths. Because SANE 
creates tokens by slicing layer weights row-wise, any variation in the layer width alters the token count and shifts the global indices for all subsequent layers. This index shift misaligns learned positional semantics and prevents the model from constructing coherent policies even within the training distribution.

On the other hand, NNiT performs very close to the training set with over $85\%$ success across all tasks. We attribute this robustness specifically to our patch-based tokenization. By decomposing weights into locally consistent patches, NNiT decouples the generative prior from the global weight space dimensions. Synthesizing a wider layer is analogous to increasing the resolution of a generated image, where the model aggregates a larger number of valid functional patches. The same idea extends naturally to depth---adding layers corresponds to appending additional architecture tokens and their associated weight blocks to the sequence, mirroring the scalability of joint audio-video generation. 

We report the full architecture performance distribution across all 80 generated policies in Appendix~\ref{app:extended_performance} and provide further analysis on a single topology in Appendix~\ref{app:single_dist}. We further verify that the pipeline transfers beyond robotic control by evaluating on MNIST classification (results summarized in Appendix~\ref{app:mnist}); we show that the NNiT attains $96.1\%$ mean test accuracy across unseen architectures.

\begin{table}[!t]
    \centering
    \caption{\textbf{Architecture Diversity \& Generalization.} Specific functional networks sampled from the generated distribution. Note that policy architectures list hidden layer widths only, omitting input and output layers. The bottom section highlights Zero-Shot Synthesis on topologies entirely unseen (Test Set) during training. Notably, NNiT achieves 98\% success on the unseen configuration $[\text{input}, 32, 16, 16, 16, \text{output}]$.}
    \label{tab:joint_generation_architecture}
    \begin{tabular}{lccc}
        \toprule
        \textbf{Policy Architecture} & \textbf{Partition} & \textbf{Return} & \textbf{Success} \\
        \midrule
        \multicolumn{4}{l}{\textit{Training (Seen) Topologies}} \\
        32, 16, 64, 32 & Seen & 39.69 & 100\% \\
        32, 64, 16, 32 & Seen & 39.01 & 100\% \\
        32, 32, 32, 16 & Seen & 38.99 & 100\% \\
        64, 64, 16, 32 & Seen & 38.81 & 98\% \\
        \midrule
        \midrule
        \multicolumn{4}{l}{\textit{Zero-Shot (Unseen) Topologies}} \\
        32, 32, 16, 64 & Unseen & 37.75 & 96\% \\
        32, 32, 32, 64 & Unseen & 37.01 & 94\% \\
        32, 16, 16, 16 & Unseen & 38.53 & 98\%\\
        16, 16, 16, 32 & Unseen & 33.80 & 94\% \\
        \bottomrule
    \end{tabular}
\end{table}

\subsection{Multimodal Joint Synthesis}
\label{sec:joint_synthesis}

A unique advantage of our patch-based representation is that it unifies the typically disjoint problems of Neural Architecture Generation (NAG)~\cite{an2023diffusionnag} and parameter generation into a single sequence modeling task. By interleaving discrete architecture tokens with continuous weight patches, NNiT learns the joint distribution $p(\mathbf{a}, \mathbf{w})$, allowing the discrete topology to serve as a causal prompt that dynamically modulates weight generation.

The results in Table~\ref{tab:joint_generation_results} quantify this capability. NNiT spontaneously generates complete network policies that achieve near-perfect success rates ($99\% \text{--} 100\%$) across tasks. This confirms that the NNiT effectively captures the conditional dependencies between topological bottlenecks and functional parameters.

Crucially, the model does not collapse to memorizing training templates. As summarized  in Table~\ref{tab:joint_generation_architecture} (Bottom), NNiTs generate functional policies for unseen configurations, such as $[\text{input}, 32, 16, 16, 16, \text{output}]$. The high success rate ($98\%$) on these unseen policies validates that the NNiTs have internalized the structural logic of neural design, extrapolating to the broader search space rather than simply recalling memorized instances. As such, we can spontaneously generate diverse, high-performing neural network policies. We report the full distribution of the generated policies in Appendix~\ref{app:joint_diversity} {and provide visual rollouts in Appendix~\ref{app:visualizations}.

\section{Limitations and Future Work}
\label{app:limitation}

Our experiments currently cover MLPs with up to four hidden layers due to computational limits. This is not a limitation of NNiT---patch tokenization supports larger widths and depths, but deeper models are harder to train as neural weights are continuous and unbounded, with high variance. Future work will adapt stabilization techniques from pixel space diffusion models~\cite{yu2025pixeldit} to better handle this dynamic range.

Structurally, framing the network depth as a temporal dimension and weight matrices as spatial features renders our approach analogous to video synthesis. This alignment allows NNiT to leverage efficiency optimizations from Video Diffusion Transformers~\cite{chen2025sana}, such as linear attention mechanisms~\cite{xie2024sana}. These techniques reduce quadratic complexity, facilitating the training of general-purpose generators capable of synthesizing billion-parameter foundation models.

Finally, NNiTs support flexible deployment for embodied AI. Rather than training separate policies for each target architecture, a single generator can be used to synthesize weights that satisfy user-specified constraints, such as a width or a compute budget.~Moreover, because NNiT conditions weight generation on discrete tokens, the same mechanism could be extended beyond architecture to other conditioning signals---such as environment or task tokens---enabling meta-learning and rapid adaptation to changing~configurations.

\section{Conclusion} 
In this work, we set out to address two barriers to generating fully-functional neural networks: the coupling of weight generation and fixed matrix dimensions, and the permutation symmetries that make MLP parameterizations inherently unaligned. The Neural Network Diffusion Transformers (NNiTs) resolve the first by replacing global vectorization with patch representation, so widening a layer corresponds to generating additional tokens rather than changing the token space. We resolve the second barrier by leveraging Graph HyperNetworks (GHNs) with a CNN decoder, which induces consistently organized weight tensors with reliable local correlations, making patch-based modeling feasible.

With this structure in place, NNiTs treat discrete architecture tokens and continuous weight patches as a single multimodal sequence, enabling both joint architecture-weight generation and conditional weight synthesis. We demonstrate that on ManiSkill3, this design yields robust zero-shot generalization to architectures and widths unseen during training, while vectorized baselines fail.

\section*{Acknowledgements}
This work is sponsored in part by the AFOSR under the award number FA9550-19-1-0169, and by the NSF under NAIAD Award 2332744 as well as the National AI Institute for Edge Computing Leveraging Next Generation Wireless Networks, Grant CNS-2112562.

\section*{Impact Statement}
This paper advances the methodology of neural network generation and weight space representation learning, contributing to the broader field of Machine Learning. There are many potential societal consequences of our work, none of which we feel must be specifically highlighted here.

\bibliography{reference}
\bibliographystyle{icml2026}

\newpage
\appendix
\onecolumn

\section{Mixture of Noise Level (MoNL) Diffusion Framework}
\label{app:monl}

We adapt the MoNL framework to a dual-modality state space consisting of discrete architectural tokens and continuous weight parameters. We employ a simplified subset of the original mixing strategies, utilizing vanilla joint diffusion and per modality (Pm). This enables simultaneous modeling of the joint synthesis distribution and the use of architecture as a clean conditional prior.

Let the state be denoted by $\mathbf{z}_0 = [\mathbf{z}^{(a)}_0; \mathbf{z}^{(w)}_0]$ where $\mathbf{z}^{(a)}_0 \in \mathbb{R}^{S \times d}$ represents the embedded architecture tokens and $\mathbf{z}^{(w)}_0 \in \mathbb{R}^{L_{seq} \times d}$ represents the projected weight patches. We employ a vectorized timestep $\mathbf{t} = (t_a, t_w)$ to independently control the noise level of each modality.

To train the model for both joint design and conditional weight generation, we sample $\mathbf{t}$ via a reference timestep $t_{\text{ref}} \sim \mathcal{U}(1, T)$ and a mode indicator $m \sim \mathcal{U}(0, 1)$. The effective timesteps are assigned as follows:
\begin{equation}
t_w = t_{\text{ref}}, \quad t_a = 
    \begin{cases} 
        t_{\text{ref}} & \text{if } m \geq 0.5 \text{ (Joint Training)}, \\
        0 & \text{if } m < 0.5 \text{ (Conditional Training)}.
    \end{cases}
\end{equation}
When $m < 0.5$, we enforce $t_a=0$, effectively conditioning the weight generation on a clean architecture embedding. To enforce the clean boundary condition, we define a target noise tensor $\tilde{\boldsymbol{\epsilon}} = [\tilde{\boldsymbol{\epsilon}}^{(a)}; \tilde{\boldsymbol{\epsilon}}^{(w)}]$ where $\tilde{\boldsymbol{\epsilon}}^{(k)} = \boldsymbol{\epsilon} \sim \mathcal{N}(0, \mathbf{I})$ if $t_k > 0$ and $\tilde{\boldsymbol{\epsilon}}^{(k)} = \mathbf{0}$ if $t_k = 0$.
We train a joint denoising network $\epsilon_\theta(x_{\mathbf{t}}, \mathbf{t})$ to predict this target noise while simultaneously learning the variance $\Sigma_\theta$. The total training objective combines Mean Squared Error (MSE) with the variational lower bound~($\mathcal{L}_{\text{vb}}$):
\begin{equation}
    \mathcal{L}_{\text{total}} = \sum_{k \in \{a, w\}} \left( \underbrace{\| \tilde{\epsilon}^{(k)} - \epsilon_\theta^{(k)}(x_{\mathbf{t}}, \mathbf{t}) \|^2}_{\text{MSE}} +  \mathcal{L}_{\text{vb}}^{(k)} \right).
\end{equation}
For the variational term, we utilize discretized Gaussian log-likelihoods for the architecture embeddings to handle their discrete structure, and standard Continuous Gaussian Log-Likelihoods for the weight parameters.


\section{Dataset Distribution Analysis}
\label{app:data_dist}

\begin{table}[!b]
    \centering
    \caption{\textbf{Quantitative Diversity Metrics.} Evaluation of 100 filtered policy weights. The high $L_2$ distances and low cosine similarities confirm that the dataset maintains significant parametric diversity despite the structural alignment.}
    \label{tab:dataset_diversity}
    \begin{tabular}{lcc}
    \toprule
    \textbf{Environment} & \textbf{Mean NN $L_2$ ($\uparrow$)} & \textbf{Mean NN Cosine ($\downarrow$)} \\
    \midrule
    PickCube-v1       & 46.94 & 0.17 \\
    PushCube-v1       & 49.66 & 0.12 \\
    StackCubeEasy-v1  & 72.79 & 0.21 \\
    \bottomrule
    \end{tabular}
\end{table}

We provide supporting visualizations and quantitative metrics to validate the structural properties proposed in Section~\ref{sec:ghn_vs_sgd}.

Figure~\ref{fig:dataset_heatmap} provides empirical evidence that the GHN's explicit locality bias projects parameters onto a structurally aligned field. The distinct vertical banding patterns observed across the 100 filtered weight samples indicate that specific spatial regions consistently encode identical functional roles across independent generations. We further note that tasks sharing identical state spaces (\texttt{PickCube-v1} and \texttt{StackCubeEasy-v1}) exhibit similar alignment signatures. This confirms that the GHN effectively propagates the fixed topological anchors of the input and output layers through the hidden topology.

We analyze the distributional properties of the weights in Table~\ref{tab:dataset_diversity} to verify that this structural alignment does not induce mode collapse. We compute the mean Nearest Neighbor (NN) metrics for the 100 independent samples of a fixed architecture. The results show high mean Euclidean distances (46.94--72.79) and low cosine similarities (0.12--0.21) across all evaluated tasks. Figure~\ref{fig:dataset_l2_cossine} complements these metrics, illustrating that the dataset retains significant parametric dispersion.


\begin{figure}[!t]
    \centering
    \includegraphics[width=0.78\linewidth]{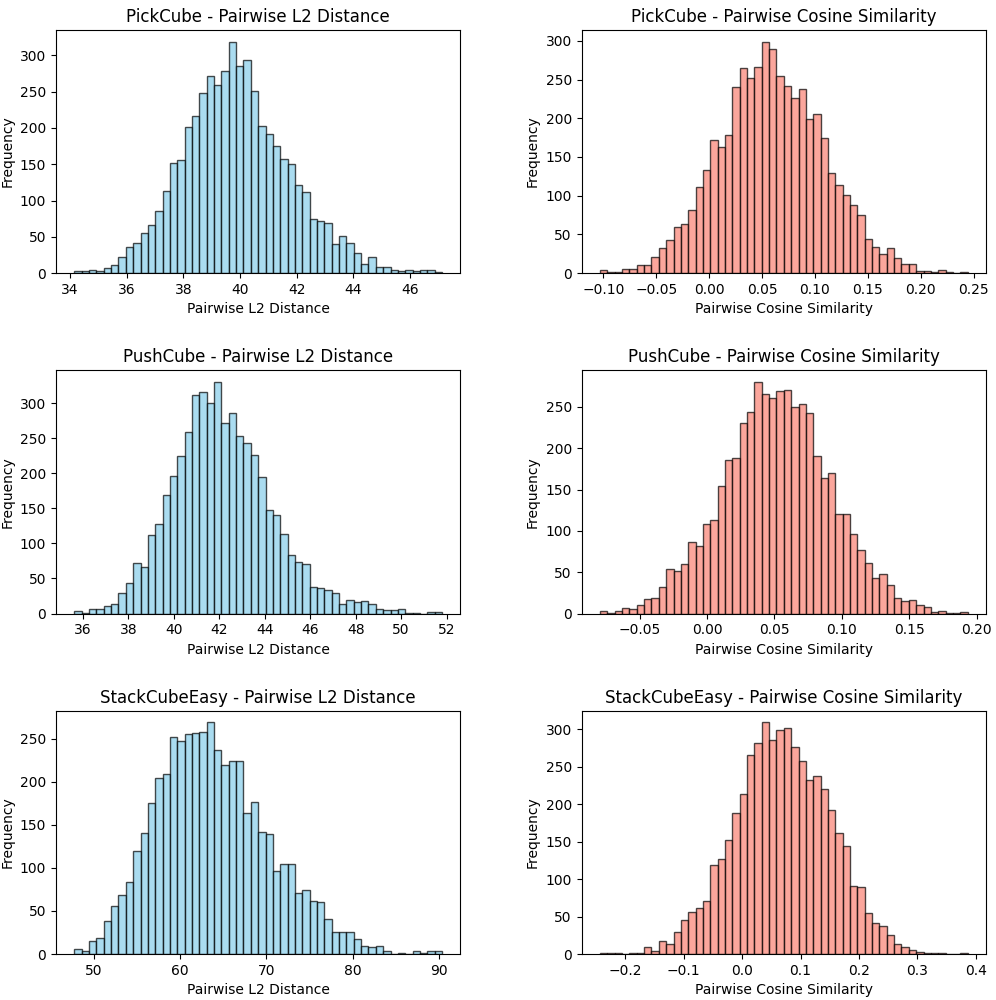}
    \caption{\textbf{Dataset Diversity Analysis.} Histograms of pairwise $L_2$ distances and Cosine Similarities across all three environments. The consistently high $L_2$ distances and low cosine similarities confirm that the structural alignment imposed by the GHN does not result in mode collapse.}
    \label{fig:dataset_l2_cossine}
\end{figure}

\begin{figure}[!t]
    \centering
    \includegraphics[width=\linewidth]{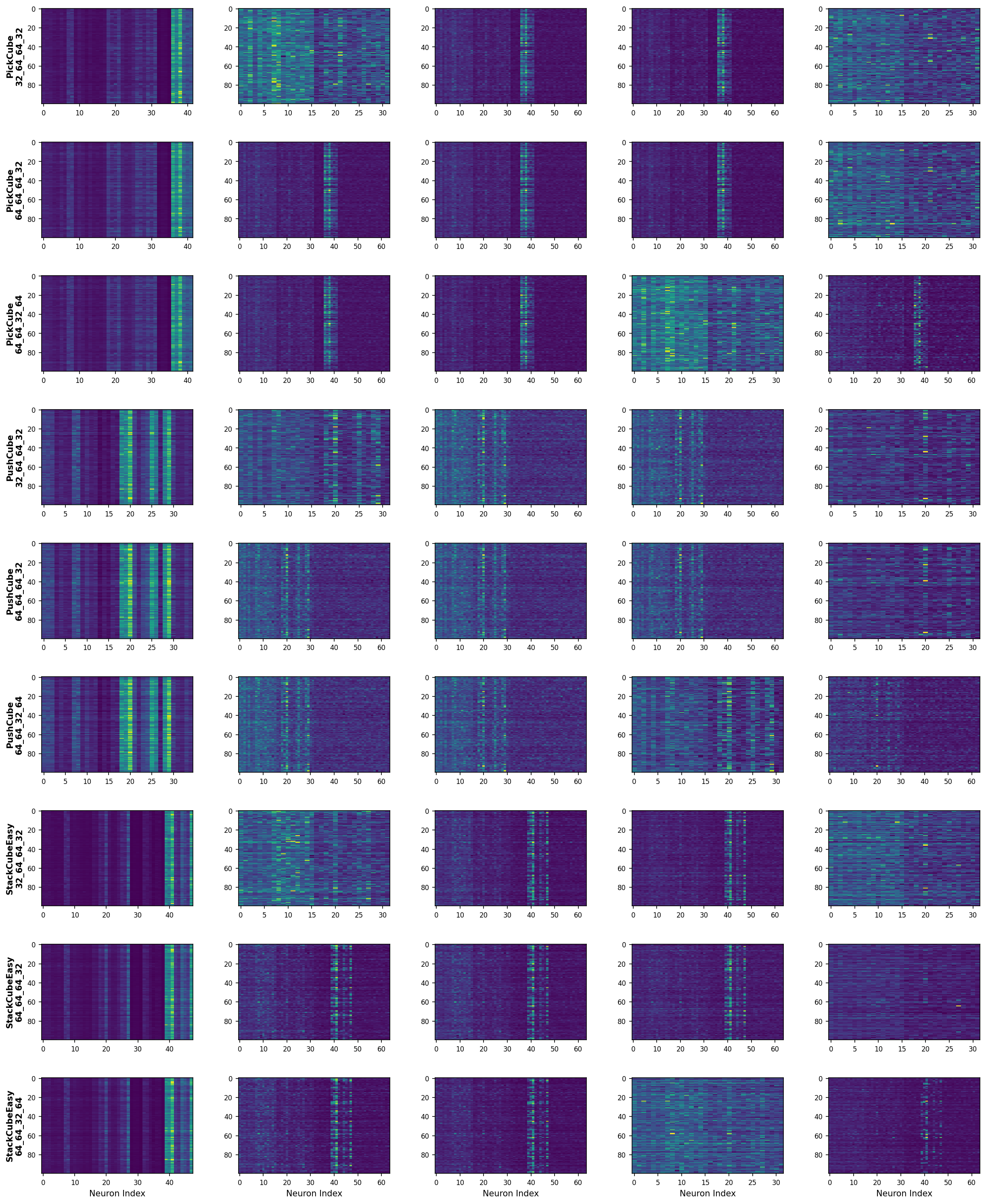}
    \caption{\textbf{Visualization of Topological Anchoring.} Heatmaps of neuron-wise weight magnitudes for 3 selected architectures across 100 filtered seeds. The vertical banding visually confirms the induction of spatial correlation, validating the premise that these weights can be treated as continuous fields.}
    \label{fig:dataset_heatmap}
\end{figure}


\section{Analyses of Structural Alignment}
\label{app:alignment_ablation}
GHN-generated weights exhibit structural alignment across seeds, whereas SGD-trained networks do not (Section~\ref{sec:ghn_vs_sgd}).  We investigate this alignment with two analyses. A permutation ablation destroys the structural alignment while preserving all weight statistics, showing that NNiT's width-agnostic generation depends on the alignment. A centered kernel alignment (CKA) analysis then confirms that the aligned weights remain functionally diverse, ruling out mode collapse.

\subsection{Representational Similarity Analysis via CKA}
\label{app:cka}
We compute pairwise CKA between the GHN-generated and SGD-trained $[256,256,256]$ MLPs on \texttt{PickCube-v1}. Lower CKA indicates greater functional diversity, so a collapsing generator would manifest as \emph{higher} CKA among GHN-generated networks.
 
\begin{table}[!t]
\centering
\small
\caption{\textbf{Representational Diversity via CKA.} Lower CKA indicates greater functional diversity between independently trained networks. GHN-generated networks exhibit lower CKA than SGD-trained networks at every layer, with the gap widening in deeper layers.}
\label{tab:cka}
\begin{tabular}{lcc}
\toprule
\textbf{CKA} & \textbf{GHN} & \textbf{SGD} \\
\midrule
Overall   & $0.849 \pm 0.088$ & $0.933 \pm 0.030$ \\
Layer 1   & $0.917 \pm 0.016$ & $0.957 \pm 0.005$ \\
Layer 2   & $0.880 \pm 0.033$ & $0.926 \pm 0.022$ \\
Layer 3   & $0.751 \pm 0.083$ & $0.918 \pm 0.037$ \\
\bottomrule
\end{tabular}
\end{table}
 
As reported in Table~\ref{tab:cka}, GHN-generated networks exhibit lower CKA than SGD-trained networks at every layer, with the gap widening in deeper layers (Layer~3: $0.751$ vs.\ $0.918$). Under mode collapse, the opposite would hold. Together with the weight-diversity metrics in Table~\ref{tab:ghn_sgd_metrics}, this confirms that the GHN supplies spatial alignment without collapsing functional diversity.
 
\subsection{Random Permutation Ablation}
\label{app:permutation_ablation}
We apply random neuron permutations to every GHN-generated MLP policy in the training corpus. This preserves the input-output function and all per-layer weight statistics, but scrambles the row-column ordering. Training NNiT on the permuted corpus collapses zero-shot performance from $78.8\%$ to $0\%$ ($0/80$ policies) on \texttt{PickCube-v1}, by the same mechanism that breaks SANE~\cite{schuerholt2024sane}.
 
The permutation also yields a controlled approximation of SGD-style weight populations. Intact GHN-generated weights exhibit a cross-seed magnitude correlation of $r \approx 0.65$, whereas both permuted GHN-generated and SGD-trained weights drop to $r \approx 0.00$. Because the permutation leaves the per-layer weight statistics unchanged, the drop reflects the loss of spatial structure rather than a change in the weight distribution.
 

\section{Dataset Generation Methodology}
\label{app:data_gen}

We construct a dataset of neural network policies using a teacher-student distillation pipeline leveraging the MLP Graph HyperNetworks (GHNs) from HyperPPO~\cite{hegde2023hyperppo}.

We target three robotic manipulation tasks from the ManiSkill3 benchmark~\cite{taomaniskill3}: \texttt{PickCube-v1}, \texttt{PushCube-v1}, and a custom variant \texttt{StackCubeEasy-v1}. We designed \texttt{StackCubeEasy-v1} to ensure target stability by instantiating the goal object as a kinematic body to remain stationary during interaction.

The ground-truth behavioral prior is derived from a single multi-task expert policy trained via Proximal Policy Optimization (PPO). This expert utilizes a 3-layer Multi-Layer Perceptron (MLP) architecture with hidden dimensions of $[256, 256, 256]$ and \texttt{Tanh} activations. The expert is trained for 10M to 50M steps across parallel environments using the standard RL baseline procedure.

To populate the architectural search space, we train an ensemble of 128 independent GHNs from scratch ($H_\phi$), initialized with distinct random seeds. This procedure ensures the final corpus captures the parametric variance inherent to the GHN optimization landscape.

We populate the final dataset by sampling from a discrete search space of 4-layer MLPs. Layer widths are drawn from the vocabulary $\{\text{input}, 16, 32, 64,\text{output}\}$. Architectures with a width of 16 at depths 3 and 4 are excluded due to insufficient capacity. This process yields a training set of 64 distinct architectures and a held-out test set of 8 unseen topologies. For each candidate, we generate weights and evaluate policy performance over $N_{eval}=64$ episodes. We retain the top-100 validated policies per architecture meeting task-specific success thresholds, resulting in a corpus of 6,400 expert policies (Table~\ref{tab:dataset_filtering}). All weights are globally normalized to a target standard deviation to align the parameter distribution for diffusion training~\cite{Peebles2022}.

\begin{table}[!t]
    \centering
    \small 
    \caption{Hyperparameters for Dataset Generation and Filtering}
    \label{tab:dataset_filtering}
    \begin{tabular}{lccc}
        \toprule
         & \multicolumn{3}{c}{\textbf{Environment Specifics}} \\
        \cmidrule(lr){2-4}
        \textbf{Parameter} & \textbf{PickCube-v1} & \textbf{PushCube-v1} & \textbf{StackCubeEasy-v1} \\
        \midrule
        \textit{GHN Training} & & & \\
        \quad Iterations & 500 & 500 & 1500 \\
        \quad Beta Decay & $0.97$ & $0.97$ & $0.98$ \\
        \quad Buffer Size & 1000000 & 1000000 & 1000000 \\
        \quad Meta Batch Size & 8 & 8 & 16 \\
        \quad Start LR & 1e-3 & 1e-3  & 1e-3  \\
        \quad End LR &  7e-4 & 7e-4 &  7e-4 \\
        
        \midrule
        \textit{Filtering Criteria} & & & \\
        \quad Min Success Rate ($\tau_{success}$) & 0.9 & 0.9 & 0.8 \\
        \quad Min Return ($\tau_{return}$) & 35.0 & 35.0 & 30.0 \\
        \quad Policies / Arch ($N_{policy}$) & 100 & 100 & 100 \\
        \bottomrule
    \end{tabular}
\end{table}

\section{Implementation Details}
\label{app:impl_details}

Specific training hyperparameters and optimization settings are detailed in Table~\ref{tab:hyperparams}. 



\begin{table}[!t]
\centering
\small
\caption{\textbf{Hyperparameter Configuration.} Training and optimization settings for the diffusion process.}
\label{tab:hyperparams}
\setlength{\tabcolsep}{5pt}
\begin{tabular}{lccc}
\toprule
\textbf{Hyperparameter} & \textbf{PickCube} & \textbf{PushCube} & \textbf{StackCubeEasy} \\
\midrule
\multicolumn{4}{l}{\textit{Diffusion Process}} \\
Timesteps ($T$) & 1000 & 1000 & 1000 \\
Noise Schedule & Linear & Linear & Linear \\
Prediction Target & Noise ($\epsilon$) & Noise ($\epsilon$) & Noise ($\epsilon$) \\
\midrule
\multicolumn{4}{l}{\textit{Optimization}} \\
Optimizer & AdamW & AdamW & AdamW \\
Weight Decay & 0.0 & 0.0 & 0.0 \\
Grad Clip & 1.0 & 1.0 & 1.0 \\
EMA Rate & 0.9999 & 0.9999 & 0.9999 \\
Precision & bfloat16 & bfloat16 & bfloat16 \\
LR Schedule & Cosine & Cosine & Cosine \\
Warmup Steps & 500 & 500 & 500 \\
Peak LR & $7 \times 10^{-5}$ & $7 \times 10^{-5}$ & $7 \times 10^{-5}$ \\
Min LR & $3 \times 10^{-5}$ & $3 \times 10^{-5}$ & $3 \times 10^{-5}$ \\
Batch Size & 16 & 16 & 16 \\
Training Epochs & 1300 & 1000 & 1300 \\
\bottomrule
\end{tabular}
\end{table}



\section{Extended Performance Analysis on Unseen Architectures}
\label{app:extended_performance}
Table~\ref{tab:conditional_generation_results} follows the standard top-$10$ protocol of \cite{liang2024make, schuerholt2024sane, soro2024diffusionbasedneuralnetworkweights}. To verify that the gap over D2NWG is distributional rather than a selection effect, we additionally report the full 80-policy distribution on the unseen architectures (Table~\ref{tab:full_distribution}).
 
\begin{table}[!t]
\centering
\small
\caption{\textbf{Full Distribution on Unseen Architectures.} Mean (over all 80 generated policies), Top-10, and Worst-Arch. NNiT's worst single architecture outperforms D2NWG's overall mean on both \texttt{PickCube-v1} and \texttt{StackCubeEasy-v1}.}
\label{tab:full_distribution}
\setlength{\tabcolsep}{6pt}
\begin{tabular}{llccc}
\toprule
\textbf{Metric} & \textbf{Method} & \textbf{PickCube-v1} & \textbf{PushCube-v1} & \textbf{StackCubeEasy-v1} \\
\midrule
\multirow{2}{*}{Mean (80 policies)}
& NNiT (Ours) & $\mathbf{78.8\%}$ & $\mathbf{74.7\%}$ & $\mathbf{54.9\%}$ \\
& D2NWG       & $15.6\%$ & $60.0\%$ & $12.2\%$ \\
\midrule
\multirow{2}{*}{Top-10}
& NNiT (Ours) & $\mathbf{98.8\%}$ & $\mathbf{100\%}$ & $\mathbf{86.4\%}$ \\
& D2NWG       & $59.4\%$ & $89.2\%$ & $41.8\%$ \\
\midrule
\multirow{2}{*}{Worst Architecture}
& NNiT (Ours) & $\mathbf{64.8\%}$ & $\mathbf{59.6\%}$ & $\mathbf{42.6\%}$ \\
& D2NWG       & $5.6\%$ & $41.2\%$ & $2.0\%$ \\
\bottomrule
\end{tabular}
\end{table}
 
NNiT's worst architecture on \texttt{PickCube-v1} ($64.8\%$) exceeds D2NWG's overall mean ($15.6\%$), so top-$10$ selection accounts for only a small share of the gap. The three tasks also vary in difficulty according to their success tolerance. \texttt{PushCube-v1} allows $0.10$\,m, \texttt{PickCube-v1} requires $0.025$\,m, and \texttt{StackCubeEasy-v1} demands $0.02$\,m. Tighter tolerances leave less margin for weight error and therefore demand more precise weight generation. D2NWG's collapse tracks this difficulty, with its mean success dropping from $63\%$ on seen architectures to $16\%$ on the unseen \texttt{PickCube-v1}, and from $58\%$ to $12\%$ on \texttt{StackCubeEasy-v1}. Even on \texttt{PushCube-v1}, the most forgiving task, D2NWG loses $21$ percentage points, indicating the benchmark is not saturated for width generalization.

\section{Single Unseen Architecture Analysis}
\label{app:single_dist}

We investigate the potential for mode collapse by generating 100 MLPs conditioned on a previously unseen topology with hidden dimensions $[16,64,64,32]$. We assess potential mode collapse by calculating the pairwise Euclidean distances and cosine similarities among the top-10 performing policies, ensuring the model preserves diversity even within optimal solutions.

Figure~\ref{fig:top10_test_arch_dist} illustrates the pairwise diversity metrics. The heatmaps show high Euclidean distances and low cosine similarities, confirming that NNiT generates distinct weights for the identical architectures. We note a single instance of redundancy, which we attribute to stochastic sampling variance given the limited scale of the training set.

\begin{figure}[H]
    \centering
    \includegraphics[width=\linewidth]{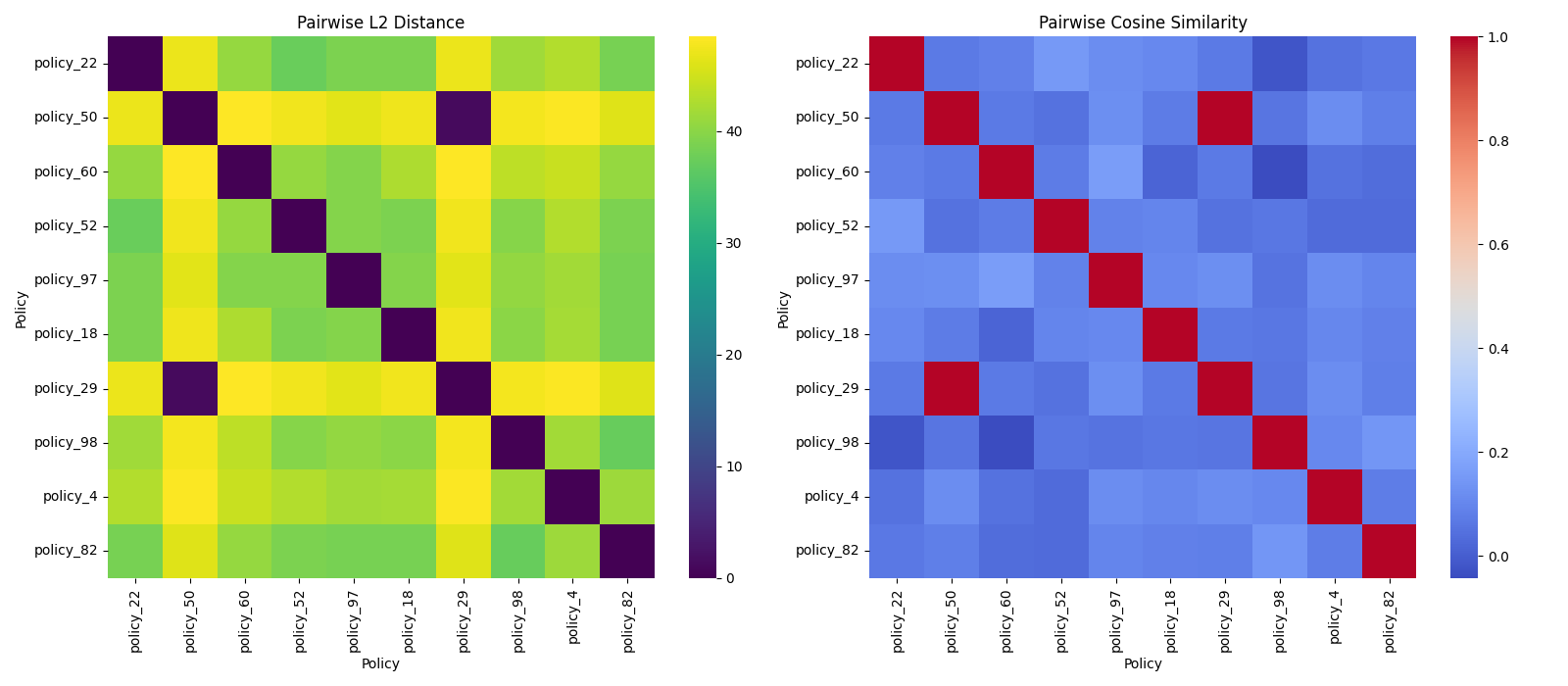}
    \caption{\textbf{Parametric Diversity Analysis.} Pairwise Euclidean distance (left) and cosine similarity (right) matrices for the top 10 generated policies. The dominance of high distances and low cosine similarities confirms that the generative model preserves variance within the parameter space, avoiding mode collapse.}
    \label{fig:top10_test_arch_dist}
\end{figure}

\section{Validation on the MNIST Benchmark}
\label{app:mnist}
MNIST model zoos are a standard benchmark for weight-space representation learning, so we additionally evaluate NNiT in this setting. We reuse the same GHN data generation, patch tokenization, width vocabulary $\mathcal{V}=\{\text{input}, 16, 32, 64, \text{output}\}$, and 4-hidden-layer MLP family used in our main experiments, changing only the task.
 
\paragraph{Dataset construction.}
MNIST's $784$-dimensional input far exceeds the input scale our setting supports, so we downsample it to $32$ dimensions via PCA fit on the training split. The GHNs are then trained on this family, and the top-$100$ validated networks per architecture are retained.
 
\paragraph{Results.}
Across $80$ generated networks on the $8$ held-out architectures used in our main experiments, NNiT attains $96.1\%$ mean test accuracy, with every network exceeding $80\%$. The pipeline therefore extends beyond robotic control, since the spatial consistency that NNiT exploits is a property of GHN-generated weight populations rather than of any task-specific~bias.

\section{Architecture Diversity under Joint Sampling}
\label{app:joint_diversity}
For joint sampling, we draw $100$ unconditional samples from $p(\mathbf{a}, \mathbf{w})$ on \texttt{PickCube-v1}. The model produces $53$ unique architectures spanning the $3^4{=}81$ width configurations, including 3 of the 8 held-out test topologies. Before selection, $97\%$ of generated policies are functional. After top-$10$ selection, $9$ distinct topologies remain with $99\%$ mean success. The diversity observed under fixed-architecture sampling therefore carries over to unconditional joint generation.

\section{Qualitative Policy Rollouts}
\label{app:visualizations}

In this section, we provide qualitative visualizations of the MLPs generated by NNiT. We visualize the execution of the generated MLPs across the three  environments. These rollouts demonstrate that the sampled MLPs successfully capture the functional logic required for delicate robotic manipulation. 


\begin{figure}[!t]
    \centering
    \includegraphics[width=\linewidth]{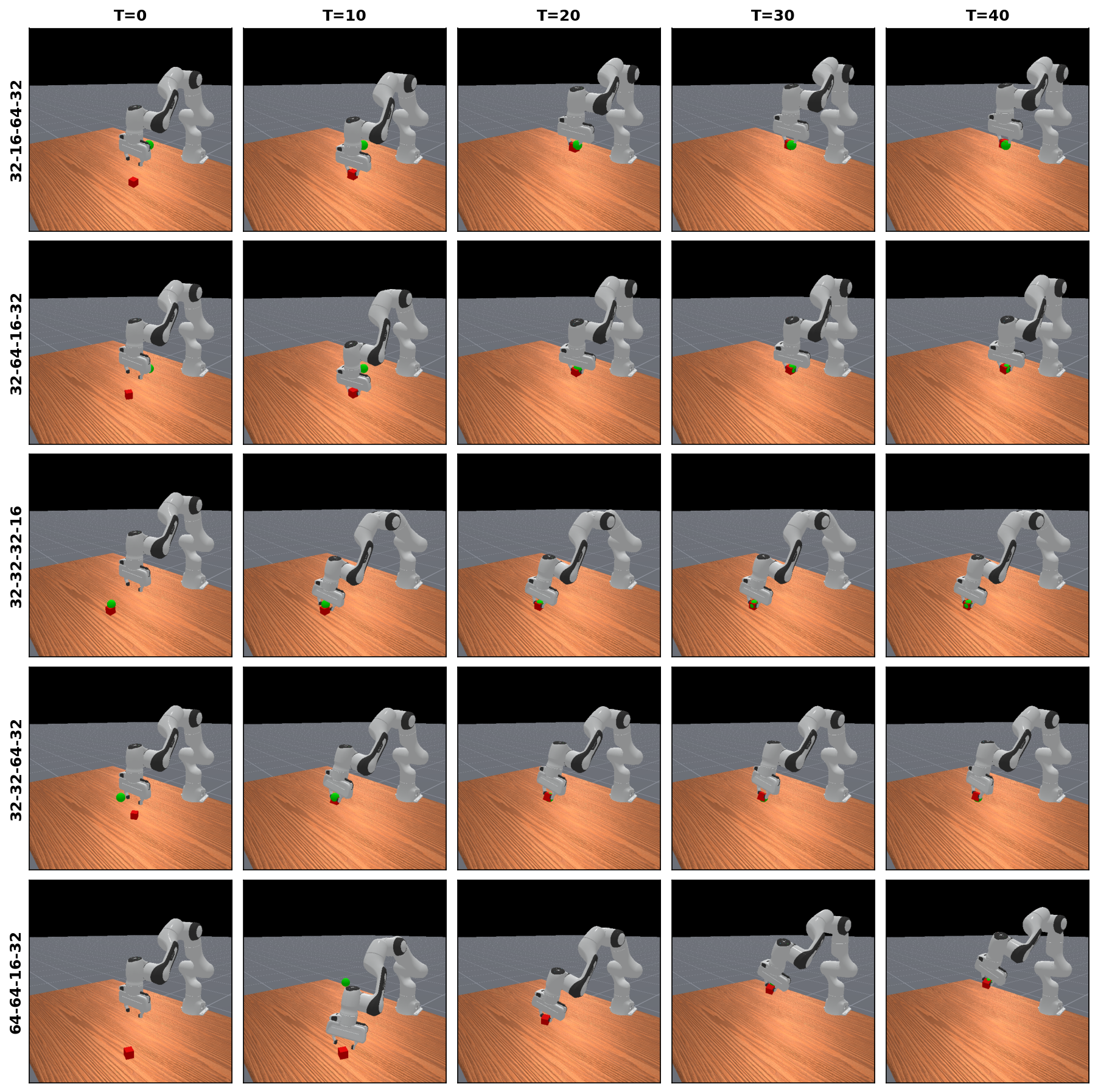}
    \caption{\textbf{PickCube-v1 Policy Rollout.} A sequential visualization of policies generated by NNiT across diverse topologies. \textbf{Left (Y-axis):} Target architecture configurations denoted by their hidden layer widths (e.g., 32-16-64-32). \textbf{Top (X-axis):} Temporal snapshots of the rollout in 10-step increments. The agent successfully navigates to the target object, executes a stable grasp, and lifts the cube to the designated goal, visually demonstrating that the sampled NNiT MLPs operate correctly.}
    \label{fig:viz_pick}
\end{figure}

\begin{figure}[!t]
    \centering
    \includegraphics[width=\linewidth]{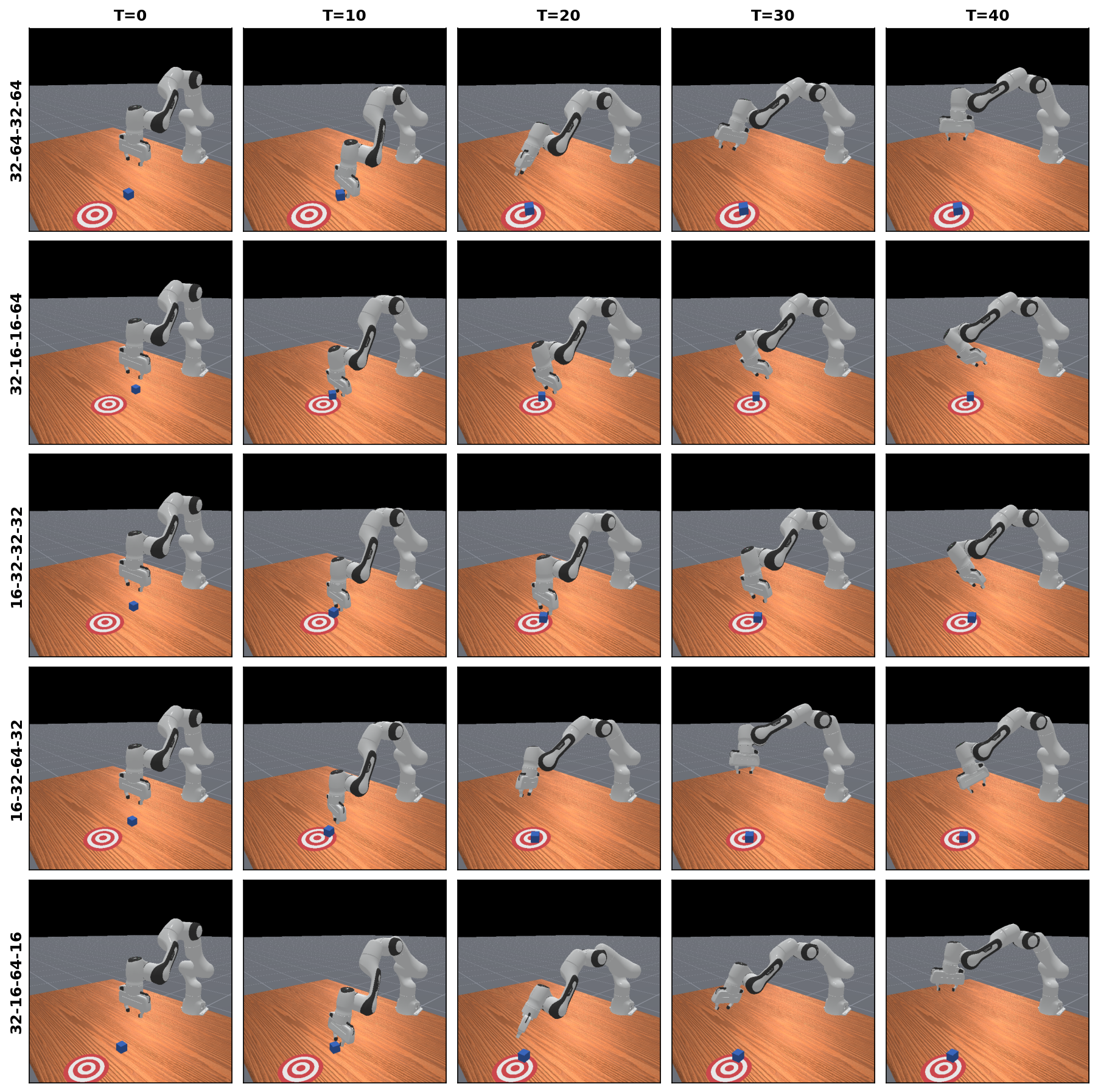}
    \caption{\textbf{PushCube-v1 Policy Rollout.} A sequential visualization of policies generated by NNiT across diverse topologies. \textbf{Left (Y-axis):} Target architecture configurations denoted by their hidden layer widths (e.g., 32-64-32-64). \textbf{Top (X-axis):} Temporal snapshots of the rollout in 10-step increments. The policy effectively coordinates the end-effector to manipulate the object, pushing it into the target goal region.}
    \label{fig:viz_push}
\end{figure}

\begin{figure}[!t]
    \centering
    \includegraphics[width=\linewidth]{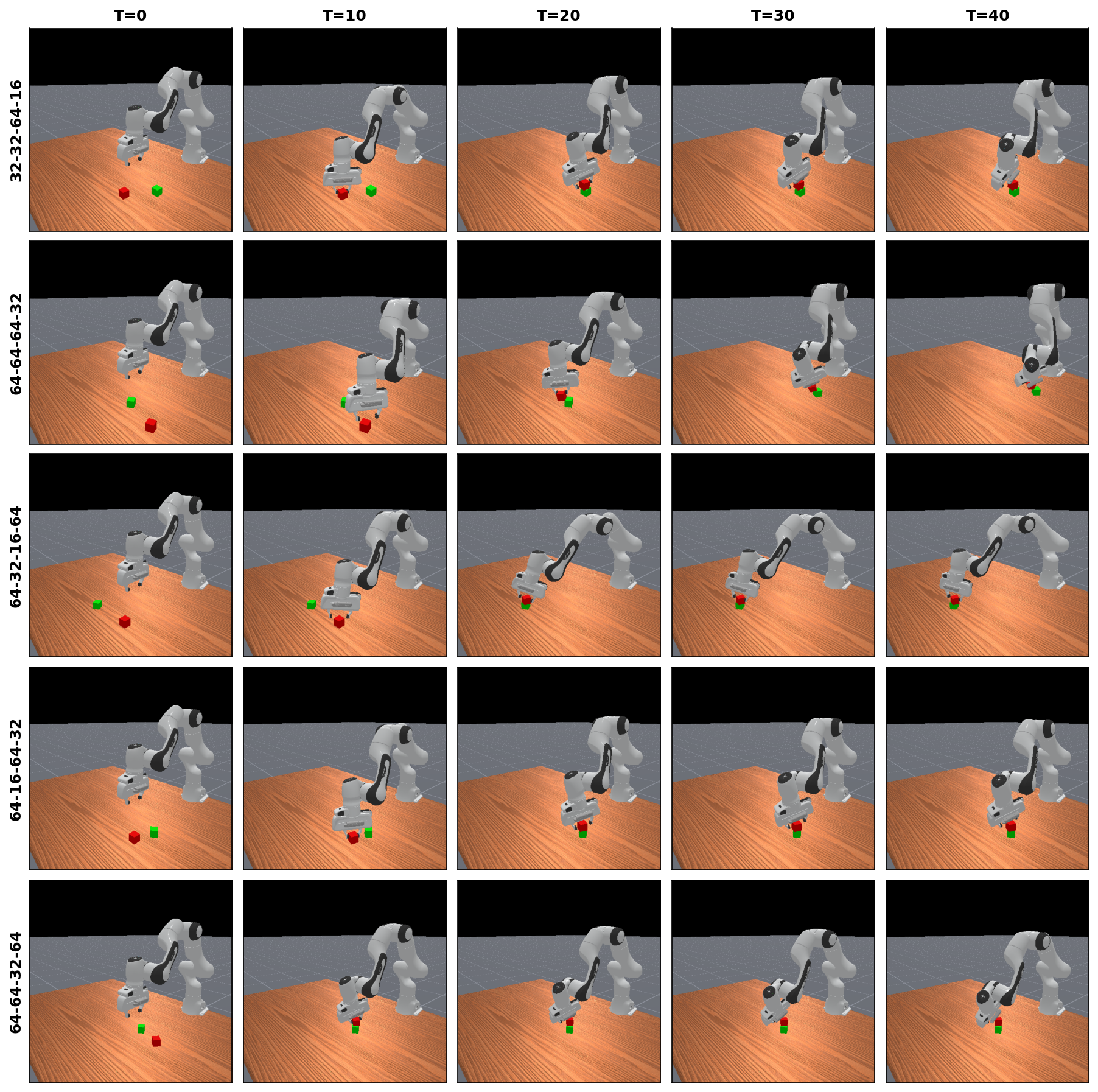}
    \caption{\textbf{StackCubeEasy-v1 Policy Rollout.} A sequential visualization of policies generated by NNiT across diverse topologies. \textbf{Left (Y-axis):} Target architecture configurations denoted by their hidden layer widths (e.g., 64-64-64-32). \textbf{Top (X-axis):} Temporal snapshots of the rollout in 10-step increments. The synthesized agent demonstrates high-precision control by successfully grasping the red cube and placing it on top of the green goal cube.}
    \label{fig:viz_stack}
\end{figure}




\end{document}